\documentclass[runningheads]{llncs}

\usepackage{graphicx}
\usepackage{amsmath}
\usepackage{amssymb}
\usepackage{arydshln}
\usepackage{xcolor,ulem}
\usepackage{comment}
\usepackage{lineno}
\usepackage{gensymb}

\begin{document}

\title{Towards Explainable Automated Neuroanatomy}

%
\author{Kui Qian$^{1,\ast}$, Litao Qiao$^1$, Beth Friedman$^2$, Edward O'Donnell$^3$,\\ David Kleinfeld$^{3,4}$ and Yoav Freund$^{2,5,\ast}$}

\authorrunning{K. Qian et al.}

\institute{University of California, San Diego \\\email{k1qian@ucsd.edu}}

\maketitle              
{\scriptsize
\noindent
$^{1}$ Department of Electrical and Computer Engineering, University of California, San Diego, La Jolla, CA 92093, USA\\
$^{2}$ Department of Computer Science and Engineering, University of California, San Diego, La Jolla, CA 92093, USA\\
$^{3}$ Department of Physics, University of California, San Diego, La Jolla, CA 92093, USA\\
$^{4}$ Department of Neurobiology, University of California, San Diego, La Jolla, CA 92093, USA\\
$^{5}$ Halıcıoğlu Data Science Institute, University of California, San Diego, La Jolla, CA 92093, USA\\
$^{\ast}$ Correspondence: yfreund@ucsd.edu, k1qian@ucsd.edu \\}

\begin{abstract}
We present a novel method for quantifying the microscopic structure of brain tissue. 
It is based on the automated recognition of interpretable features obtained by analyzing the shapes of cells. This contrasts with prevailing methods of brain anatomical analysis in two ways. 
First, contemporary methods use gray-scale values derived from smoothed version of the anatomical images, which dissipated valuable information from the texture of the images. 
Second, contemporary analysis uses the output of black-box Convolutional Neural Networks, while our system makes decisions based on interpretable features obtained by analyzing the shapes of individual cells. 
An important benefit of this open-box approach is that the anatomist can understand and correct the decisions made by the computer.
Our proposed system can accurately localize and identify existing brain structures. 
This can be used to align and coregistar brains and will facilitate connectomic studies for reverse engineering of brain circuitry.

\keywords{Explainable ML \and Brain texture \and Computational anatomy}
\end{abstract}

\section{Introduction}

One of the first steps in brain analysis is to answer the ``where'' question. 
To answer this question the anatomist typically relies on brain cytoarchitecture, namely, the spatial organization of neural elements. 
However, manual labeling of the brain structures is a labor-intensive task. 
Typically, identifying and marking the boundaries of 40 standard landmarks in a single brain takes a trained anatomist many weeks of work.

\begin{figure}[t]
\includegraphics[width=\textwidth]{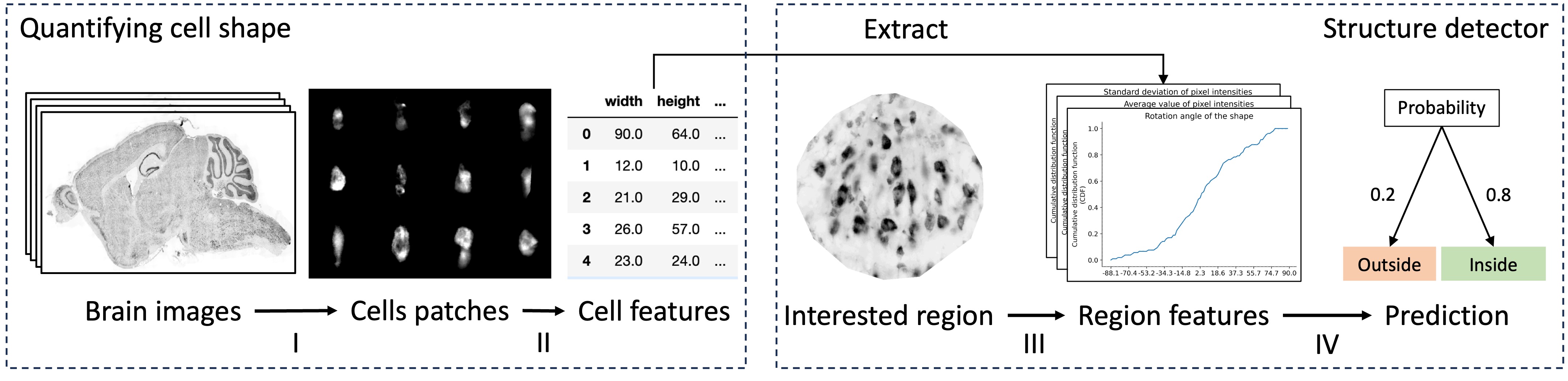}
\caption{    
    The pipelines of our proposed system, where the figures indicate the outputs of the stages and Roman letters enumerate the procedures. 
    I: OpenCV for cell segmentations. 
    II: K-means, Diffusion Mapping, and feature space alignment for extracting cell features.
    III: Cumulative CDFs for generating regional features.
    IV: XGBoost for the classification of the region. 
}
\label{fig:pipeline}
\end{figure}

Based on \cite{chen2019active}, our new work introduces a machine learning-based approach to automate the identification of structures within high-resolution mouse brain images (Figure~\ref{fig:pipeline}). 
The method in \cite{chen2019active} introduced the use of high resolution texture of brain tissue to identify different brain regions. 
The process of identification relied on a pre-trained convolutional neural network (CNN) that processes pixel-level features. This ``black box'' approach leads to inherently uninterpretable results.
Our new system utilizes interpretable cell shape features for structure detection. 
Our underlying assumption is that the distribution of the cell shapes inside and outside the structures should be significantly different, which is analogous to the criteria used by anatomists for structure identification~\cite{Braitenberg1977OnTT}.
By focusing on individual cells as the primary unit of analysis, our method not only makes self-explainable decisions for the anatomists but also maintains robustness across different staining procedures and imaging techniques. 
We demonstrate that the features we compute for cell shapes can be used as inputs to structure detectors of different brain regions.

Of interest, a recent independent study \cite{tajduhar2023InterpretableML} also uses cell shapes to build an interpretable machine learning method for cortical cytoarchitecture analysis.
This study solved a classification challenge of detection of human cortex laminae using a neuron-centric approach.
Here we apply a neuron-centric approach to detect the more general case of non-laminated geometrically diverse spatial distributions of neurons. Such distributions are particularly prevalent in the mouse brainstem, a region that is typically very challenging to map.
Our method relies on two innovative approaches.
Firstly, on top of manually designed features, our approach utilizes unsupervised learning to extract nuanced and interpretable features that more accurately represent the diversity of cell shapes. 
By not relying solely on predefined features, our system can uncover patterns that might otherwise be overlooked.
Secondly, we incorporate regional features that summarize the statistical attributes of cell populations, rather than individual cells alone. 
These features capture the collective properties of cell groups, such as density and orientation distributions, to provide a robust framework for our detector. 
This approach enables the system to recognize and classify structures even when individual cell shapes are ambiguous or when cells exhibit subtle differences that are only discernible as a population. 
The integration of these advanced regional features significantly enhances the detector's robustness, making it more resilient to variations in staining methods and imaging modalities.

\section{Methods}
\label{sec:methods}

\subsubsection{Quantifying cell shapes}

Since our method is based on identifying the shapes of the cells inside and outside the interested structure, the first part of our method is to generate \textit{cell features} to quantify the shapes of all cells in a single brain. 
Our method starts with cell segmentations from all brain images using OpenCV~\cite{opencv_library}, a library of computer vision tools that includes adaptive thresholding and connected components to isolate cell images from the noisy images. 
Then all cell images are zero-padded to create pre-defined uniformly sized cell patches for efficient handling of the subsequent processing steps.
To efficiently represent each cell patch, we employ a dimensionality reduction technique via Diffusion Mapping (DM)~\cite{belkin2003,coifman2005geometric}. 
This creates a continuous non-linear mapping from single-cell patches into $m$-dimensional feature vectors that represent the $m$ most significant eigenvectors of the Laplace-Beltrami operator.
In our case, we choose $m=10$ for all brains, which significantly reduces the dimensionality from the vast number of original pixel values to a more manageable feature space that best quantifies the cell shapes.
However, as we typically extract tens of millions of cells from each brain, the dataset used to train DM is too large to fit in the computer memory. 
We thus used a streaming implementation of the K-means algorithm~\cite{arthur2007k} as an efficient way to create a small number of representative cell patches as the training set for the DM algorithm (supplementary materials).

\begin{figure}[t]
\includegraphics[width=\textwidth]{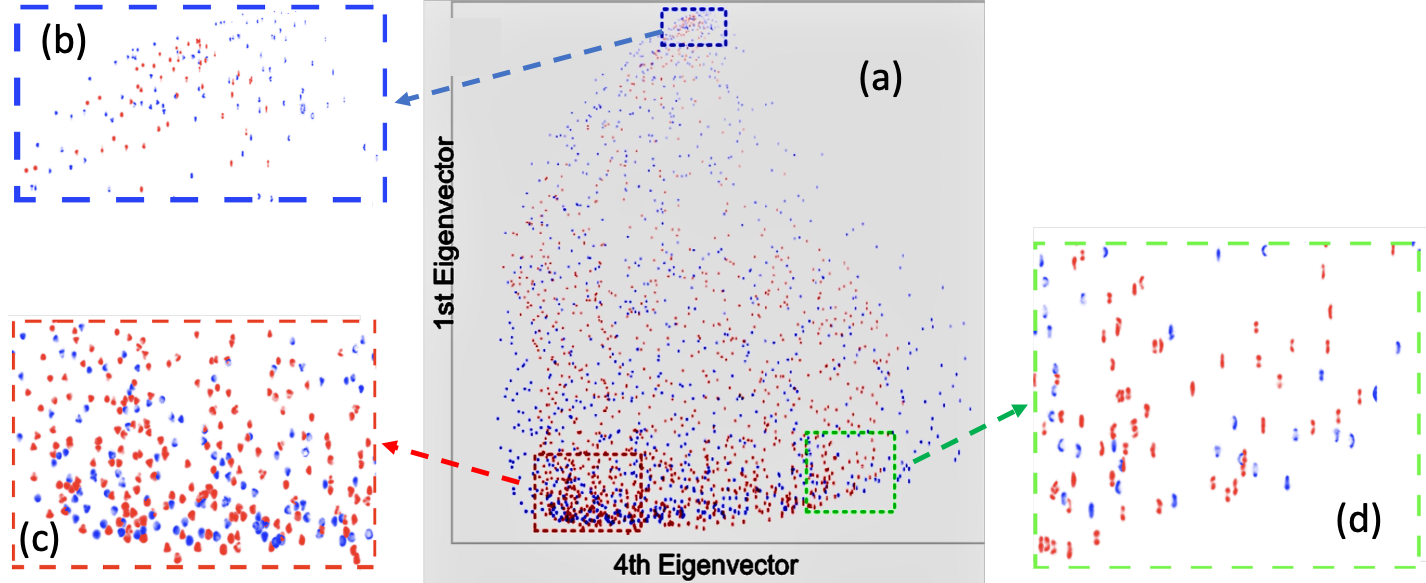}
\caption{
    (a) Visualization of patch clouds from two brains with different modalities on the 1st and 4th fixed features.
    Each ``dot'' in the image is a cell patch. 
    The red patches are from a brain that is stained with thionin and imaged using brightfield, and the blue patches are from a brain that is stained with NeuroTrace blue and imaged using fluorescence.
    (b) The region of very small cells. 
    (c) The region of large and round cells.
    (d) The region of thin cells.
}
\label{fig:diffusionmap}
\end{figure}

Conceptually, DM transforms cell patches into a 10-dimensional feature space, which allows the visualization of these patches by treating their features as coordinates within this space. 
By projecting cell patches in the same brain onto a 2D plane from this space, we create a ``patch cloud'' that represents cell shape diversity along the chosen two dimensions. 
The visualization of the patch clouds shows that the cell features learned through DM capture the cell shapes in a visually explainable way. 
Furthermore, after visually analyzing the patch clouds generated from different brains, we found that these patch clouds share similar shapes, despite the fact that the brain images are obtained using different stains (thionin vs. NeuroTrace blue) and imaging techniques (brightfield vs. fluorescent).
In particular, the clouds from the different brains only differ by the orders of the axes or the orientations of the shapes when they are visualized in the 2D planes.
This motivates the idea that the cell features from any brain can be aligned to a fixed set of features using a simple affine transformation. 
By adopting the cell features from the selected brain images as the reference, we formulated a root mean squared (RMS) optimization procedure (supplementary materials) that has a closed-form solution to formulate the affine transformation matrix.
For example, Figure~\ref{fig:diffusionmap} shows the two patch clouds projected onto the subspace formed by the 1st and 4th eigenvectors.  
Each cloud consists of the cell patches obtained using the K-means algorithm that are representatives of the cells in two different brains that are stained using thionin and NeuroTrace blue, respectively. 
The patch clouds are visualized by projection onto the subspace formed by the 1st and 4th eigenvectors. 
The clouds overlap almost perfectly (Figure~\ref{fig:diffusionmap}a), which confirms the effectiveness of the affine transformation for aligning cell features from different brains.
Further, the cell patches with similar shapes are grouped together in different regions (Figure~\ref{fig:diffusionmap}b\textendash d), which implies that the 4th eigenvector on the x-axis describes the aspect ratio of the cell shape and the 1st eigenvector on the y-axis gives the sizes of the cells.

In addition to the 10 cell features identified through the DM technique, we incorporate an additional 10 manually designed features (supplementary materials) that explicitly describe the shape of the cells.
These supplementary features are directly extracted from the images, without undergoing a learning process. 
Finally, our feature vector for each cell encompasses a total of 20 attributes. 
To facilitate the extraction of cell features for groups of cells in subsequent steps of our method, we will establish a database that stores the feature vectors for all cells within the brain.
It is important to highlight that this feature extraction process is based on unsupervised learning algorithms. 
This can be performed even without manual annotations of the structures by experts in anatomy.

\subsubsection{Structure detection}

\begin{figure}[t]
\includegraphics[width=\textwidth]{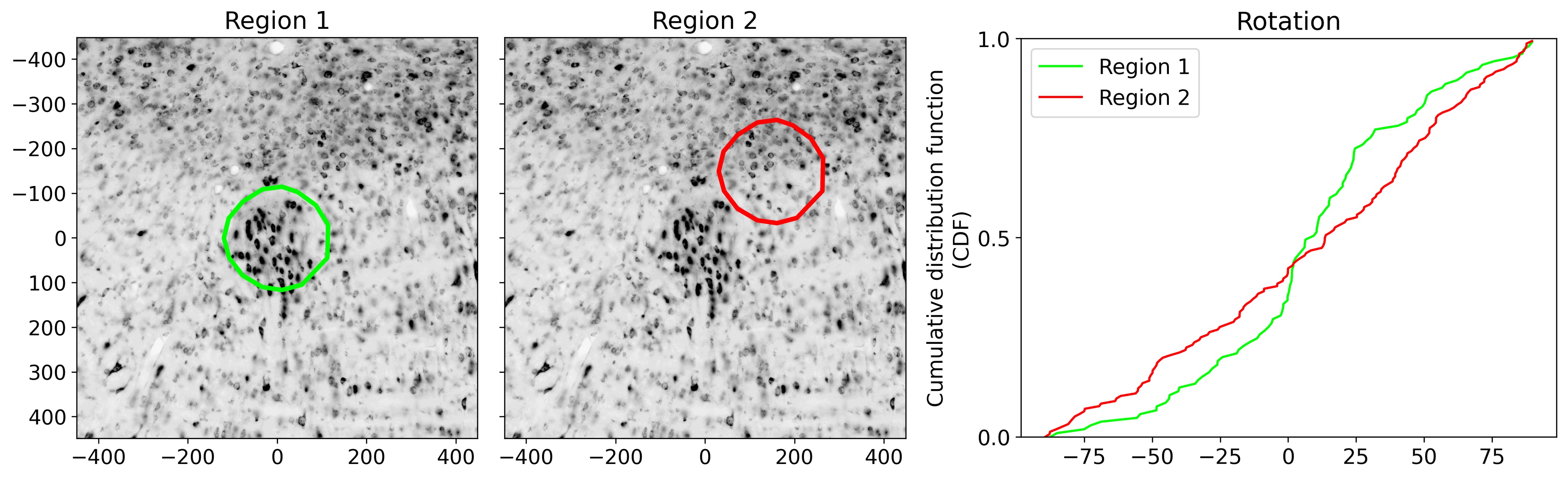}
\caption{
    Cell shape distributions for two regions near the Abducens Nucleus (6N).
    The left image highlights a region that roughly corresponds to the 6N structure (circled in green), while the right image shows a region outside the structure (circled in red). 
    The CDF graph represents the rotation feature of cells, with the green curve depicting the CDF for cells within region 1 and the red curve for those within region 2. 
}
\label{fig:cdfs}
\end{figure}

One challenge of using the neuron-centric approach for the task of structure detection lies in the fact that the single-cell shape does not possess the distinctive properties to be classified as either inside or outside a target structure. 
Since cytoarchitecture is a property of collections of cells, a well-trained anatomist usually determines the boundary of the structure based on the distribution of a group of cells in a small region.
This insight motivates us to develop our structure detection system to classify image regions rather than individual cells, which are characterized using \textit{regional features} that represent the cell shape distribution in a specific region. 
Also, analyzing regions instead of individual cells can greatly reduce the sensitivity of the analysis to segmentation errors. 
By leveraging the 20 cell features designed for characterizing individual cell shapes, we can describe a region that contains a collection of cells through the distributions of these features, which are represented by their empirical cumulative distribution functions (CDFs).
Therefore, the two regions have different characteristic cell shapes if their CDFs are different (Figure~\ref{fig:cdfs}).

To compute the region feature for a region containing a group of cells, we first query the cell feature database to retrieve features for all cells within the region, followed by generating cumulative CDFs for these cell features. 
Each region's feature CDF is then discretized into 99 points, resulting in a comprehensive region feature vector of $20 \times 99 = 1980$ dimensions.
Then we append two additional features: cell density per unit area and the area ratio covered by cells, which results in a final region feature vector with 1982 elements. 
Finally, our structure detection model takes the vector of the regional features and predicts the likelihood of the region belonging to a specific structure.
Our model employs XGBoost~\cite{chen2016xgboost}, which is a supervised learning algorithm that is particularly suitable for our case.
XGBoost not only inherits Adaboost's resistance to overfitting but also is interpretable in the sense that feature importance can be readily derived from the trained model which is critical for anatomists to understand the decision of the model.
Given that all our brains have 26 annotated structures, we opted for 26 binary classifiers rather than a single multi-class classifier, as this allows for more precise and focused detection of each specific structure.
\section{Results}
\label{sec:results}

\subsubsection{Classifier accuracy}

We evaluate our method by comparing the predictive performance of our structure detector with the CNN texture classifier introduced in \cite{chen2019active}, both quantitively and qualitatively.
The quantitive evaluation of our method was conducted using the same three human-annotated brain image stacks captured from sagittal views, stained with thionin, and scanned at a resolution of $0.5\mu m$ using brightfield imaging~\cite{chen2019active}. 
For the CNN texture classifiers, we utilized the pre-existing models from \cite{chen2019active} without further training. 
Unlike the CNN approach, whose input image dimension is limited to $224 \times 224$ pixels, our method can process regions of any shape. 
However, to ensure a fair comparison, we adopted the same preprocessing procedures as \cite{chen2019active} to prepare the training set for our method and the test set for both methods. 
In particular, all images from the three brains were split into small image tiles of size $224 \times 224$ pixels using the sliding window technique, which will be used as the input regions for our method. 
Given that we are training 26 binary classifiers corresponding to each of the 26 structures, a training image tile is labeled as positive for a specific classifier if more than half of the tile is within the anatomists' defined structure boundary and is labeled as negative for that classifier otherwise. 
The image tiles from two of the three brain images were used as the training set, while the third one was used as the test set for performance evaluation.

\begin{figure}[t]
\includegraphics[width=\textwidth]{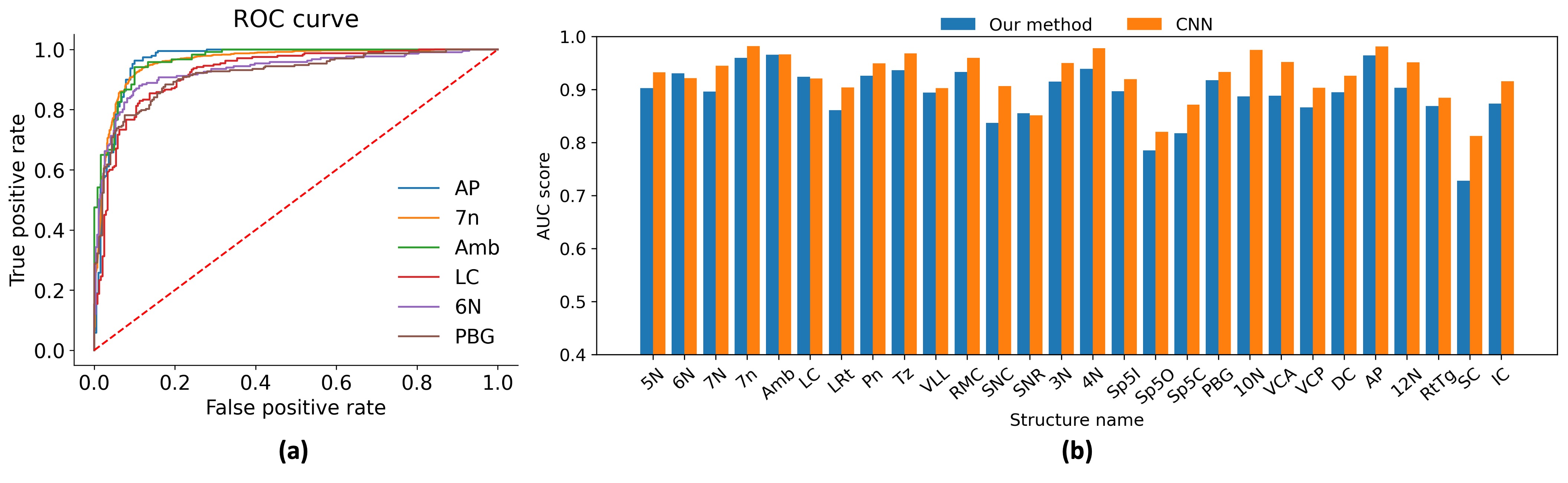}
\caption{
    Comparison of ROC AUC scores for detecting different structures. 
    Please see \cite{chen2019active} for the list of all 26 structures and their abbreviations. 
}
\label{fig:roc_auc}
\end{figure}

The performance of both methods was assessed using the Area Under the Receiver Operating Characteristic Curve (ROC AUC) metric.
The ROC curves for six chosen structures using our method are shown in Figure~\ref{fig:roc_auc}a, while Figure~\ref{fig:roc_auc}b displays the scores of both methods across all 26 structures.
The ROC AUC scores for our method consistently demonstrate high predictive performance, with the lowest scores surpassing 0.75 and the highest nearing 1.0. 
Thus our method reliably differentiates between the various structures. 
The average ROC AUC score for our method is 0.892, which, while slightly lower than the CNN's average of 0.924, still represents a robust performance. 

\subsubsection{Probability map}

\begin{figure}[t]
\includegraphics[width=\textwidth]{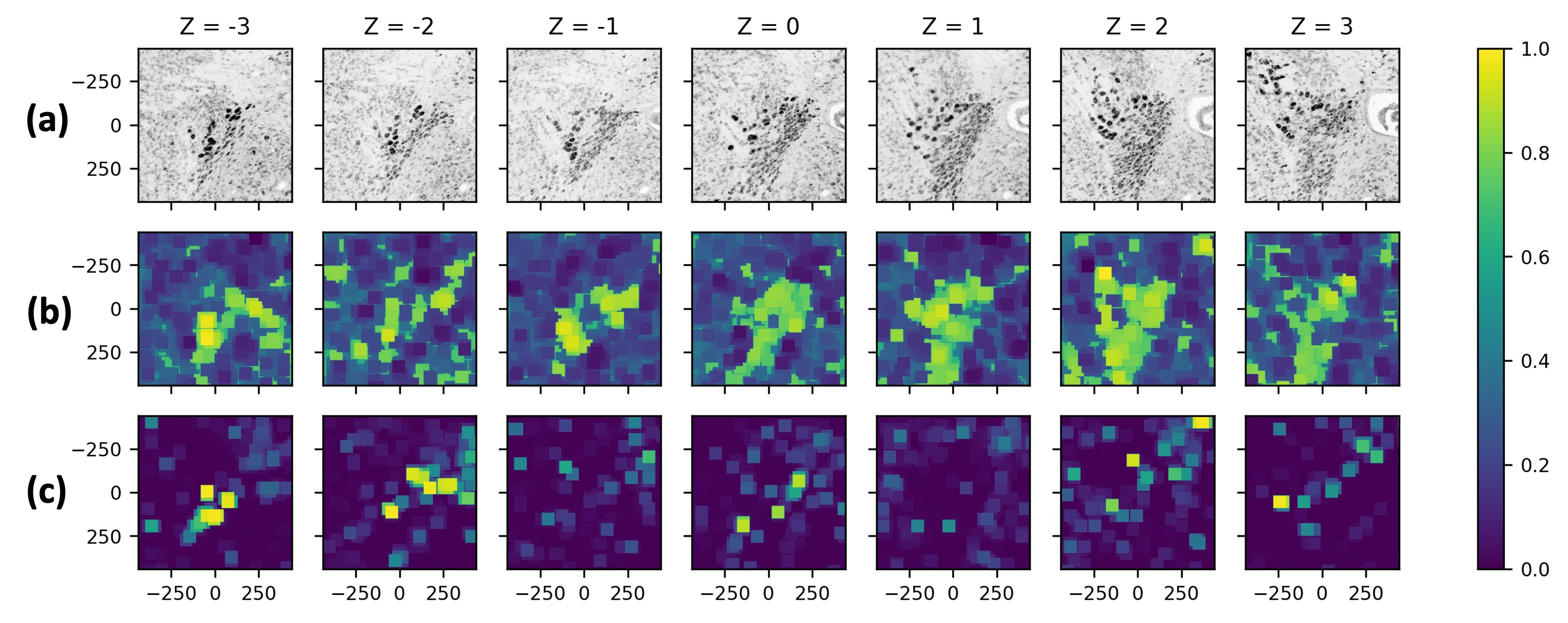}
\caption{
    Comparison of structure identification across different sections using probability maps. 
    (a) Original image patches of the LC with cells delineated in black. 
    (b) Probability maps for each of the sections of our method.
    (c) Probability maps of the CNN method.
    Bright and dark colors represent high and low probabilities of being inside the structure, respectively. 
}
\label{fig:prob-map}
\end{figure}

The unique advantage of our method is that our model can be directly applied to images that have different textures from the training images without retraining. 
In the following experiment, we used the same classifiers used in the first experiment, which were trained on thionin-stained brain images, to produce probability maps for brain images derived from a brain stained with NeuroTrace blue. 
This staining method yields brain images with pixel intensities markedly different from those stained with thionin.
We followed the same step used in \cite{chen2019active} to generate the probability maps for both methods, wherein the output of the classifier determines the probability of each image tile.
Because of the dearth of annotations of structures for this brain, we will evaluate the probability maps from both methods qualitatively through visual inspection.
In Figure~\ref{fig:prob-map}a, we show 7 different image sections of the left locus coeruleus (LC), where we can clearly see the structure in each image section defined by the black and grey cells.
Our method yields high-probability regions that are closely aligned with the actual distribution of cells, as shown in the original image patches (Figure~\ref{fig:prob-map}b). 
This alignment is visually more precise in the probability maps of our method than in the CNN probability maps (Figure~\ref{fig:prob-map}c).
Note the regions of high probability derived using the CNN method do not accurately overlay the cell-marked areas. 
This observation underscores the robustness of our method in generating structure-specific probability maps across different staining modalities, ensuring its utility without the necessity of retraining the model for each new staining technique.

\subsubsection{Understanding the detector}

Utilizing XGBoost as our classification algorithm offers the distinct advantage of providing feature importance metrics for each input variable. 
This is particularly useful given that our cell and regional features are intrinsically interpretable, which allows us to give anatomists a visual explanation for the model's decisions. 
Specifically, we can trace back to the original images and highlight cells that satisfy a feature deemed critical by XGBoost in identifying a particular structure.

\begin{figure}[t]
\includegraphics[width=\textwidth]{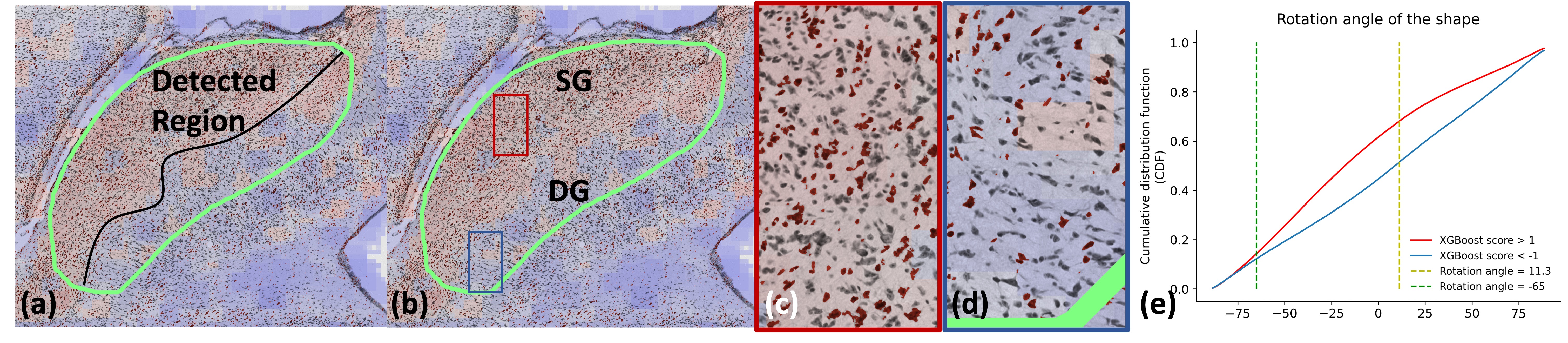}
\caption{
    Explanation of the SC structure detection.
    (a) The result of our detector (green contour), where there is a clear boundary between the recognized regions (red) and unrecognized regions (blue).
    (b) The SG and DG gray layers within the SC, which roughly correspond to the red and blue regions predicted by our detector.
    (c) and (d) A closer view of the cellular orientation that our detector uses for its classification shows a concentration of cells oriented from the lower left to the upper right in the red region, in contrast to the sparsity of such cells in the blue region.
    (e) The CDFs of the cell rotation angles for cells in high-score and low-score regions, where the cells with angles between $-65 \degree$ and $11.3 \degree$ are highlighted in (c) for a visual explanation. 
}
\label{fig:understanding}
\end{figure}

We now consider explanations of our automated method. In Figure~\ref{fig:understanding}a the green outline labels the superior colliculus (SC) as identified by anatomists. In contrast, the areas our detector identifies as SC are shaded in red, which is suboptimal as there is a large area of the unrecognized region (blue color) inside the structure. 
Intriguingly, this behavior is understandable as the discernible boundary marked by the black line made by our detector aligns with known biological layers within SC: the superficial (SG) and deep (DG) grey~\cite{franklin2019paxinos} (Figure~\ref{fig:understanding}b).
We can understand why our method makes such decisions by highlighting the cells that satisfy the features deemed important by the XGBoost model.
In this case, the XGBoost model put the highest importance on the feature ``rotation\textendash 11.3'', which represents the cumulative probability that a cell's rotation angle is less than $11.3 \degree$. 
By highlighting the cells whose rotation angles are in the range between $-65 \degree$ and $11.3 \degree$ in dark red, we find that the highlighted cells predominantly populate the red regions, while being scarcely present in the blue regions (Figure~\ref{fig:understanding}c). 
Thus, a biological rationale for our detector's partitioning is primarily based on the density of the cells oriented from the lower left to the upper right. 
The disparities in the empirical CDFs for the rotation angle of cells illustrate a clear statistical distinction between the cell orientations in the regions identified with high ($> 1$) and low ($< -1$) XGBoost scores (Figure~\ref{fig:understanding}e).
It is worth noting that we deliberately omit the cells with angles less than $-65 \degree$ for visual clarity, since they contribute very little to the XGboost decision process because of the similar densities of such cells in both highlighted regions; see CDFs in Figure~\ref{fig:understanding}d. 

\section{Conclusion}
\label{sec:conclusion}

Our method introduces significant contributions to computational neuroanatomy: 
(1) an unsupervised learning procedure for efficiently extracting quantifiable cell shape features, 
(2) region features that encapsulate the statistical properties of cell populations, 
and (3) the deployment of supervised learning, specifically XGBoost, for robust structure detection. 
By focusing on cell shapes and statistical properties, our approach achieves high accuracy and interpretability, aligning closely with the criteria used by anatomists. 
Future directions include refining segmentation techniques, exploring advanced unsupervised learning models for feature extraction, extending the method's applicability to other species or brain regions, and integrating with neuroanatomical databases to uncover new brain structures.

\bibliographystyle{splncs04}
\bibliography{main}





\title{Supplementary Materials for ``Towards Explainable Automated Neuroanatomy''}
\author{Anonymous}
\institute{Anonymous Organizations \\\email{**@****.***}}

\maketitle

\subsubsection{Details about OpenCV procedure} 

    For cell segmentation from brain section images, we used the adaptive thresholding technique provided by OpenCV. This process, specifically through the \texttt{cv.adaptiveThreshold} function, computes the threshold for a pixel based on a small region around it. The parameters used in this function were as follows:
    \vspace{-5pt}
    \begin{verbatim}
    thresholdType: cv2.THRESH_BINARY,
    adaptiveMethod: cv.ADAPTIVE_THRESH_GAUSSIAN_C,
    blockSize: 101 (chosen based on the typical cell size),
    C: -12 (selected manually based on segmentation quality).
    \end{verbatim}
    \vspace{-25pt}

\subsubsection{Details about KMeans}

    We applied the Kmeans clustering algorithm on an extensive dataset of ten million cell patches extracted from a single brain's segmentation to find around one thousand representative cell patches. The Kmeans procedure included two key steps: initialization with Kmeans++ and subsequent refinement. Kmeans++ was used to select 2,000 initial cluster centroids. Then these centroids were used to aggregate similar cell patches into clusters. The mean of each cluster was computed as the final set of representative cell patches. Clusters containing fewer than 5 samples were excluded. Figure~\ref{fig:kmeans} shows samples of the original cell images and of representative cell patches selected using K-means after normalizing rotation.

\subsubsection{Details about Diffusion map}
    
    We used the public implementation of DM~\footnote{\tt https://github.com/DiffusionMapsAcademics/pyDiffMap}. The parameters used in this function were as follows:
    \vspace{-5pt}
    \begin{verbatim}
    n_evecs – Number of diffusion map eigenvectors: 100,
    k – Number of nearest neighbors to construct the kernel: 100,
    epsilon: 5000 (chosen based on the typical cell size),
    alpha: 1.0, neighbor_params: {'n_jobs': -1, 'algorithm': 'ball_tree'}.
    \end{verbatim}
    \vspace{-15pt}

\subsubsection{Details about RMS-based optimization}

    The mathematical formulation of finding a linear transformation between two different diffusion maps can be described as follows: Given a set of vector pairs $(a_1,b_1),\ldots,(a_n,b_n)$ where each of the vectors $a_i,b_i$ are in a $d$ dimensional space $R^d$,  find an offset vector $\mu \in r^d$ and a linear transformation $M$ which is a $d \times d$ matrix so that the following cost function is minimized:
    
    \begin{equation}
    \label{cost_function}
        cost(\mu, M) = \frac{1}{n}\sum_{i=1}^n||\mu + Ma_i - b_i||_2^2
    \end{equation}
    
    We first compute the hessian matrix of ~\ref{cost_function} with respect to $\mu$ and $M$ to see if the problem can be directly solved.
    \begin{equation}
    \label{hessian_1}
        \frac{\partial^2 cost(\mu, M)}{\partial \mu \partial \mu^T} = 2I \rightarrow p.d
    \end{equation}
    \begin{equation}
    \label{hessian_2}
        \frac{\partial^2 cost(\mu, M)}{\partial M \partial M^T} = \frac{2}{n}\sum_{i=1}^na_ia_i^T \rightarrow p.s.d
    \end{equation}
    
    The results show that both hessian matrices are Positive Semi-definite matrices and thus we can find the solution by setting derivatives to 0. We have:
    \begin{equation}
        (\frac{1}{n}\sum_{i=1}^na_ia_i^T - \frac{1}{n}\sum_{i=1}^na_i\times \frac{1}{n}\sum_{i=1}^na_i^T)M^T = \frac{1}{n}\sum_{i=1}^na_ib_i^T - \frac{1}{n}\sum_{i=1}^na_i\times \frac{1}{n}\sum_{i=1}^nb_i^T
    \end{equation}

\subsubsection{Details about 10 manually designed features} 

The 10 manually designed features encompass the following: width, height, and area of a cell; rotation angle along with its confidence level; mean and standard deviation of pixel intensities within the cell image; the size of the cell patch, and standard deviation of horizontal and vertical coordinates of all cell pixels in the cell patch.

\subsubsection{Details about XGBoost classifiers }

    The parameters used to train our XGBoost classifiers were as follows:
    \vspace{-5pt}
    \begin{verbatim}
    params: {'max_depth':3, 'eta': 0.2, 
            'objective':'binary:logistic', 'num_class':1},
    num_boost_round: 100.
    \end{verbatim}

\begin{figure}[t]
\includegraphics[width=\textwidth]{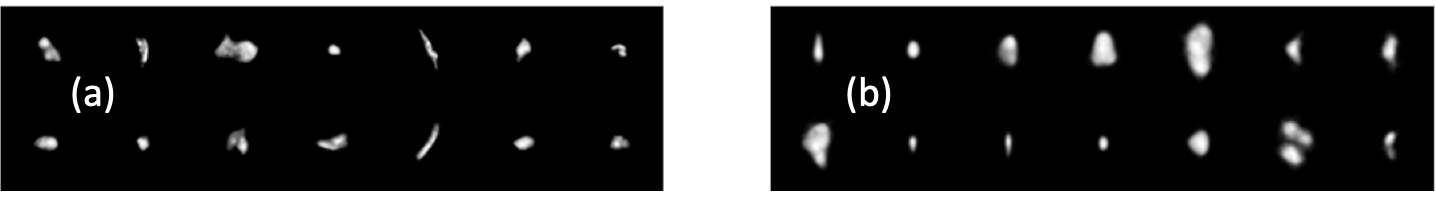}
\caption{
    {\bf (a)} Original cell patches, {\bf (b)} Cell patches selected using K-means after normalizing rotation.
}
\label{fig:kmeans}
\end{figure}

\end{document}